\documentclass{exam}
\usepackage{subfigure}
\usepackage{authblk}
\usepackage{amsmath,amsfonts}
\usepackage{algorithmic}
\usepackage{algorithm}
\usepackage{array}
\usepackage[caption=false,font=normalsize,labelfont=sf,textfont=sf]{subfig}
\usepackage{textcomp}
\usepackage{stfloats}
\usepackage{url}
\usepackage{verbatim}
\usepackage{graphicx}
\usepackage{cite}
\usepackage{enumitem}
\usepackage{caption}
\usepackage{subcaption}

\title{\textbf{Multimodal Data Integration for Oncology in the Era of Deep Neural Networks: A Review}}

\author[a]{Asim Waqas\footnote{Corresponding author: asim.waqas@moffitt.org}\footnote{Equal contribution}}
\author[a]{Aakash Tripathi${^\dag}$}
\author[b]{Ravi P. Ramachandran}
\author[a]{Paul Stewart}
\author[a]{Ghulam Rasool}

\affil[a]{Department of Machine Learning, Moffitt Cancer Center, Tampa, Florida, USA.}
\affil[b]{Department of Electrical and Computer Engineering, Rowan University, Glassboro, New Jersey, USA.}

\date{March 2024}

\begin{document}
\maketitle

\begin{abstract}
Cancer research encompasses data across various scales, modalities, and resolutions, from screening and diagnostic imaging to digitized histopathology slides to various types of molecular data and clinical records. The integration of these diverse data types for personalized cancer care and predictive modeling holds the promise of enhancing the accuracy and reliability of cancer screening, diagnosis, and treatment. Traditional analytical methods, which often focus on isolated or unimodal information, fall short of capturing the complex and heterogeneous nature of cancer data. The advent of deep neural networks has spurred the development of sophisticated multimodal data fusion techniques capable of extracting and synthesizing information from disparate sources. Among these, Graph Neural Networks (GNNs) and Transformers have emerged as powerful tools for multimodal learning, demonstrating significant success. This review delves into the recent advancements in GNNs and Transformers for the fusion of multimodal data in oncology, spotlighting key studies and their pivotal findings. We explore the foundational principles of multimodal learning and discuss the unique challenges it faces, such as data heterogeneity and integration complexities, alongside the opportunities it presents for a more nuanced and comprehensive understanding of cancer. By surveying the landscape of multimodal data integration in oncology, our goal is to underline the transformative potential of multimodal GNNs and Transformers. These models are poised to revolutionize cancer prevention, early detection, and treatment strategies by fostering informed, personalized oncology practices. By addressing both the technological advancements and the methodological innovations, we aim to chart a course for future research in this promising field. Our review highlights not only the current state of multimodal modeling applications in oncology but also sets the stage for their evolution, encouraging further exploration and development in personalized cancer care.
\end{abstract}

Cancer represents a significant global health challenge, characterized by the uncontrolled growth of abnormal cells, leading to millions of deaths annually. In 2023, the United States had around 1.9 million new cancer diagnoses, with cancer being the second leading cause of death and anticipated to result in approximately 1670 deaths daily \cite{ACAstats}. However, advancements in oncology research hold the promise of preventing nearly 42\% of these cases through early detection and lifestyle modifications. The complexity of cancer, involving intricate changes at both the microscopic and macroscopic levels, requires innovative approaches to its understanding and management. In recent years, the application of machine learning (ML) techniques, especially deep learning (DL), has emerged as a transformative force in oncology. DL employs deep neural networks to analyze vast datasets, offering unprecedented insights into cancer's development and progression \cite{ccalicskan2023ai, chen2023trends, siam2023multimodal, muhammad7artificial, talebi2024machine}. This approach has led to the development of computer-aided diagnostic systems capable of detecting and classifying cancerous tissues in medical images, such as mammograms and MRI scans, with increasing accuracy. Beyond imaging, DL also plays a crucial role in analyzing molecular data, aiding in the prediction of treatment responses, and the identification of new biomarkers \cite{varlamova2024machine, Khan2023, Muhammad2023, dera2021premium, dera2019extended, waqas2021brain, barhoumi2023efficient}. As the volume of oncology data continues to grow, DL stands at the forefront of this field, enhancing our understanding of cancer, improving diagnostic precision, predicting clinical outcomes, and paving the way for innovative treatments. This review explores the latest advancements in DL applications within oncology, highlighting its potential to revolutionize cancer care \cite{ghaffari2023facts, CAD, tripathi2024building, Ibrahim2022}.

\begin{figure}[ht!]
    \centering
    \includegraphics[width=\columnwidth]{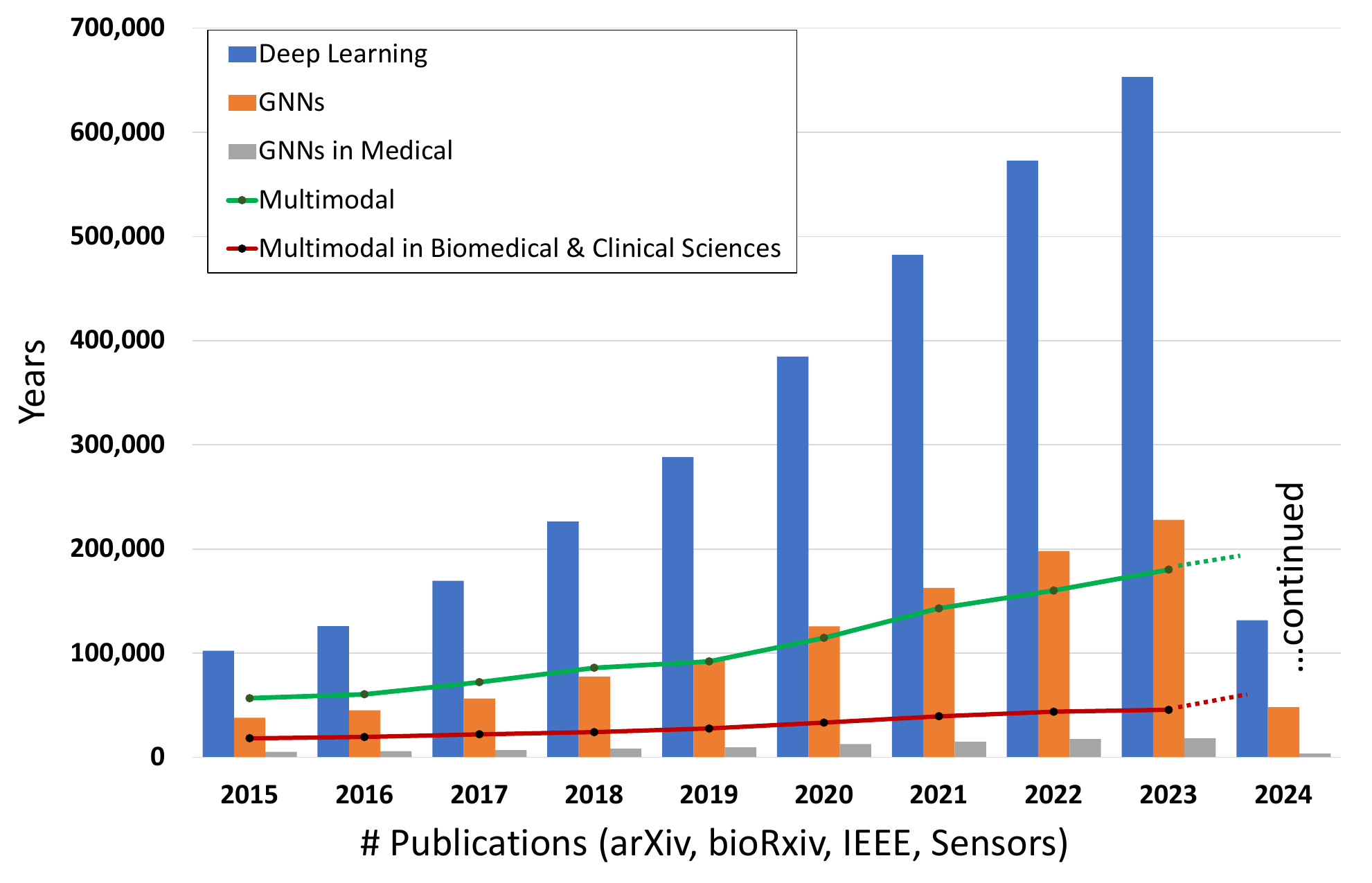}
    \caption{Number of publications involving DL, GNNs, GNNs in the medical domain, overall multimodal and multimodal in biomedical and clinical sciences in the period 2015-2024 \cite{dimensions}.}
    \label{fig:all_publications}
\end{figure}

Multimodal Learning (MML) enhances task accuracy and reliability by leveraging information from various data sources or modalities \cite{huang2021what}. This approach has witnessed a surge in popularity, as indicated by the growing body of MML-related publications (see Figure ~\ref{fig:all_publications}). By facilitating the fusion of multimodal data, such as radiological images, digitized pathology slides, molecular data, and electronic health records (EHR), MML offers a richer understanding of complex problems \cite{tripathi2024multimodal}. It enables the extraction and integration of relevant features that might be overlooked when analyzing data modalities separately. Recent advancements in MML, powered by Deep Neural Networks (DNNs), have shown remarkable capability in learning from diverse data sources, including computer vision (CV) and natural language processing (NLP) \cite{openai2023gpt4, bommasani2022opportunities}. Prominent multimodal foundation models such as Contrastive Language-Image Pretraining (CLIP) and Generative Pretraining Transformer (GPT-4) by OpenAI have set new benchmarks in the field \cite{CLIP, openai2023gpt4}. Additionally, the Foundational Language And Vision Alignment Model (FLAVA) represents another significant stride, merging vision and language representation learning to facilitate multimodal reasoning \cite{FLAVA}. Within the realm of oncology, innovative applications of MML are emerging. The RadGenNets model, for instance, integrates clinical and genomics data with Positron Emission Tomography (PET) scans and gene mutation data, employing a combination of Convolutional Neural Networks (CNNs) and Dense Neural Networks to predict gene mutations in Non-small cell lung cancer (NSCLC) patients \cite{RadGenNets}. Moreover, GNNs and Transformers are being explored for a variety of oncology-related tasks, such as tumor classification \cite{MMLtumorclassification}, prognosis prediction \cite{MMLprognosis}, and assessing treatment response \cite{MMLtreatment}.

Recent literature has seen an influx of survey and review articles exploring MML \cite{boehm2021harnessing, MLwithTransformersAsurvey, baltruvsaitis2018multimodal, ektefaie2023multimodal, hartsock2024vision}. These works have provided valuable insights into the evolving landscape of MML, charting key trends and challenges within the field. Despite this growing body of knowledge, there remains a notable gap in the literature regarding the application of advanced multimodal DL models, such as Graph Neural Networks (GNNs) and Transformers, in the domain of oncology. Our article aims to fill this gap by offering the following contributions:

\begin{enumerate}
    \item \emph{Identifying large-scale MML approaches in oncology}. We provide an overview of the state-of-the-art MML with a special focus on GNNs and Transformers for multimodal data fusion in oncology.
 
    \item \emph{Highlighting the challenges and limitations of MML in oncology data fusion}. We discuss the challenges and limitations of implementing multimodal data-fusion models in oncology, including the need for large datasets, the complexity of integrating diverse data types, data alignment, and missing data modalities and samples.

    \item \emph{Providing a taxonomy for describing multimodal architectures}. We present a comprehensive taxonomy for describing MML architectures, including both traditional ML and DL, to facilitate future research in this area.
    
    \item \emph{Identifying future directions for multimodal data fusion in oncology}. We identify GNNs and Transformers as potential solutions for comprehensive multimodal integration and present the associated challenges.
    
\end{enumerate}

By addressing these aspects, our article seeks to advance the understanding of MML's potential in oncology, paving the way for innovative solutions that could revolutionize cancer diagnosis and treatment through comprehensive data integration.

Our paper is organized as follows. Section \ref{Fundamentals of Multimodal Learning} covers the fundamentals of MML, including data modalities, taxonomy, data fusion stages, and neural network architectures. Section \ref{GNNs in Multimodal Learning} focuses on GNNs in MML, explaining graph data, learning on graphs, architectures, and applications to unimodal and multimodal oncology datasets. Section \ref{Transformers in Multi-modal Learning} discusses Transformers in MML, including architecture, multimodal Transformers, applications to oncology datasets, and methods of fusing data modalities. Section \ref{Challenges and Opportunities} highlights challenges in MML, such as data availability, alignment, generalization, missing data, explainability, and others. Section \ref{Multimodal Oncology Data Sources} provides information on data sources. Finally, we conclude by emphasizing the promise of integrating data across modalities and the need for scalable DL frameworks with desirable properties.

\section{Fundamentals of Multimodal Learning (MML)} \label{Fundamentals of Multimodal Learning}

\begin{figure}
     \centering
     \begin{subfigure}
         \centering
        \includegraphics[width=\linewidth]{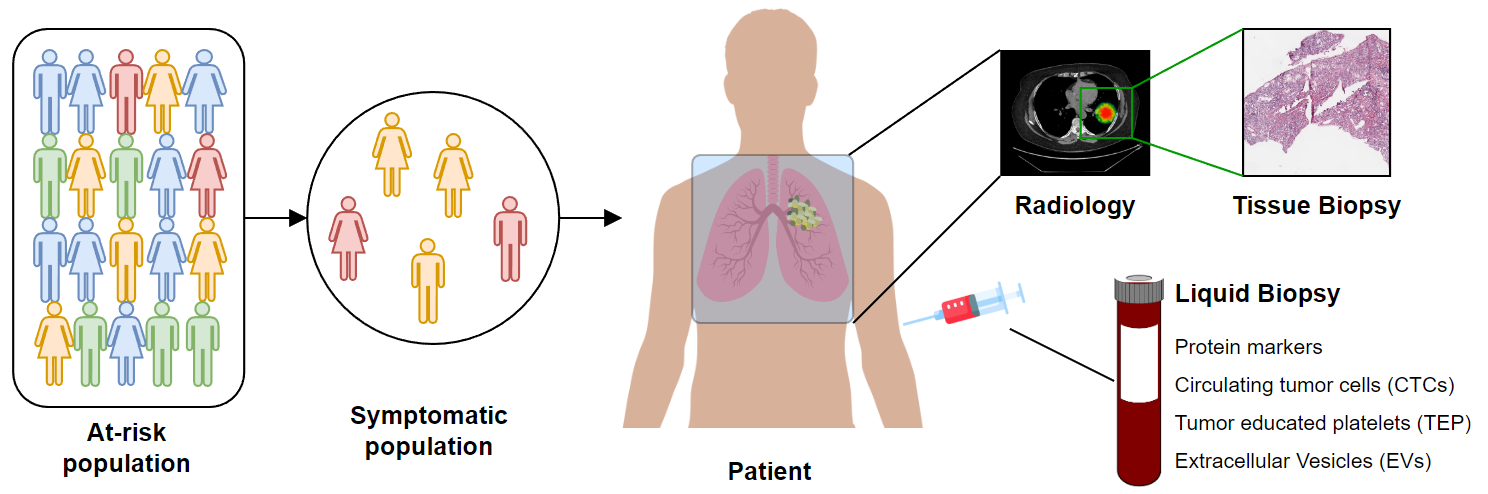}
        \caption{An oerview of data collected from population to a tissue.}
        \label{fig:patient_data_modal}
     \end{subfigure}
     \hfill
     \begin{subfigure}
         \centering
        \includegraphics[width=\linewidth]{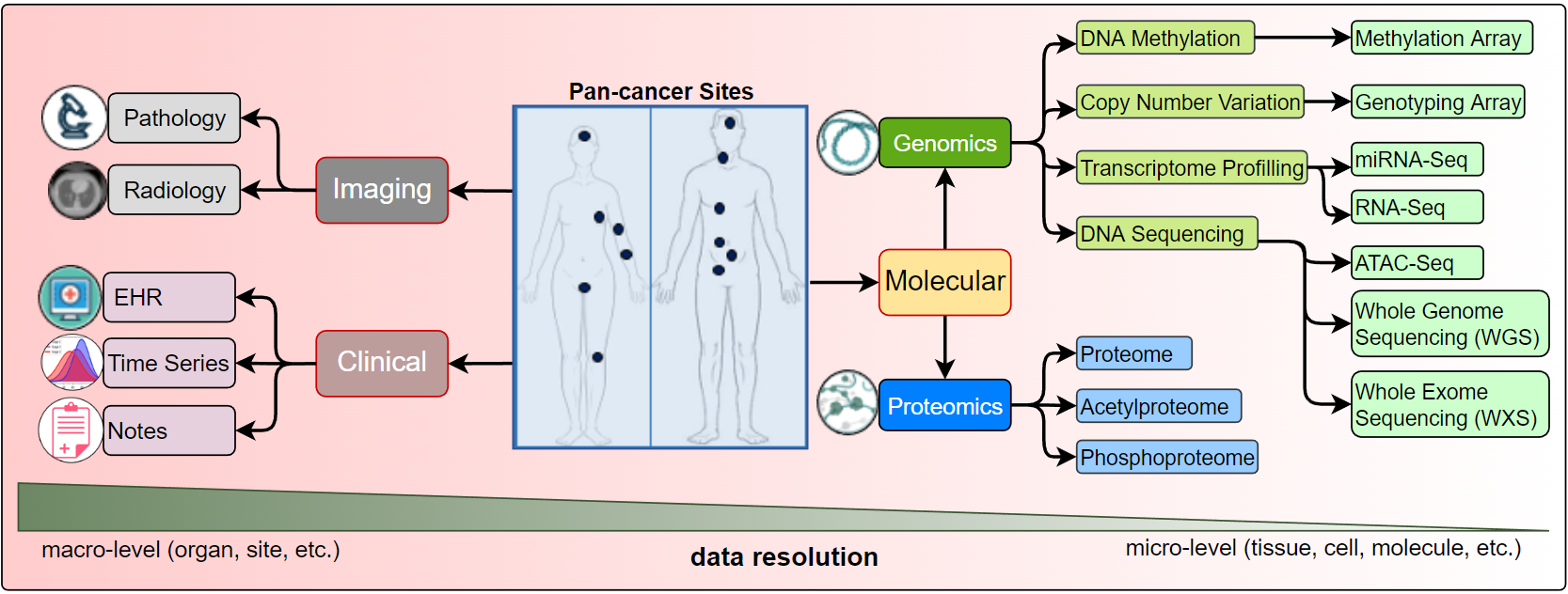}
        \caption{Detailed look into data modalities acquired for cancer care.} \label{fig:onco_modalities}
     \end{subfigure}
        \caption{We present various data modalities that capture specific aspects of cancer at different scales. For example, radiological images capture organ or sub-organ level abnormalities, while tissue analysis may provide changes in the cellular structure and morphology. On the other hand, various molecular data types may provide insights into genetic mutations and epigenetic changes.}
        \label{fig:three graphs}
\end{figure}

\subsection{Data Modalities in Oncology} \label{Data Modalities in Oncology}
A data \emph{modality} represents the expression of an entity or a particular form of sensory perception, such as the characters' visual actions, sounds of spoken dialogues, or the background music \cite{MultimodalClassification}. A collective notion of these modalities is called \emph{multi-modality} \cite{baltruvsaitis2018multimodal}.  Traditional data analysis and ML methods to study cancer data use single data modalities (e.g., EHR~\cite{miotto2016deep}, radiology~\cite{waqas2021brain}, pathology \cite{litjens2017survey}, or molecular, including genomics \cite{angermueller2017deepcpg}, transcriptomics \cite{yousefi2017predicting}, proteomics \cite{wang2017accurate}, etc.). However, the data is inherently multimodal, as it includes information from multiple sources or modalities that are related in many ways. Figure~\ref{fig:patient_data_modal} provides a view of multiple modalities of cancer at various scales, from the population level to single-cell analysis. Oncology data can be broadly classified into 3 categories: clinical, molecular, and imaging, where each category provides complementary information about the patient's disease. Figure~\ref{fig:onco_modalities} highlights different clinical, molecular, and imaging modalities. Multimodal analysis endeavors to gain holistic insights into the disease process using multimodal data.


    \subsubsection{Molecular Data}
    Molecular data modalities provide information about the underlying genetic changes and alterations in the cancer cells \cite{TherapeuticAdvancesinOncology}. Efforts toward integrating molecular data resulted in the \emph{multi-omics} research field \cite{waqas2024senmo}. Two principal areas of molecular analysis in oncology are proteomics and genomics. \emph{Proteomics} is the study of proteins and their changes in response to cancer, and it provides information about the biological processes taking place in cancer cells. \emph{Genomics} is the study of the entire genome of cancer cells, including changes in DNA sequence, gene expression, and structural variations \cite{boehm2021harnessing}. Other molecular modalities include transcriptomics, pathomics, radiomics and their combinations, radiogenomics, and proteogenomics. Many publicly available datasets provide access to molecular data, including the Proteomics Data Commons for proteomics data and the Genome Data Commons for genetic data~\cite{PDC, GDC}. 

    \subsubsection{Imaging Data}
    Imaging modalities play a crucial role in diagnosing and monitoring cancer. The imaging category can be divided into 2 main categories: (1) radiological imaging and (2) digitized histopathology slides, referred to as Whole Slide Imaging (WSI). \emph{Radiological} imaging encompasses various techniques such as X-rays, CT scans, MRI, PET, and others, which provide information about the location and extent of cancer within the body. These images can be used to determine the size and shape of a tumor, monitor its growth, and assess the effectiveness of treatments. \emph{Histopathological} imaging is the examination of tissue samples obtained through biopsy or surgery \cite{Molecularimaginginoncology, waqas2023revolutionizing}. Digitized slides, saved as WSIs, provide detailed information about the micro-structural changes in cancer cells and can be used to diagnose cancer and determine its subtype. 

    \subsubsection{Clinical Data}
    Clinical data provides information about the patient's medical history, physical examination, and laboratory results, saved in the patient's electronic health records (EHR) at the clinic. EHR consists of digital records of a patient's health information stored in a centralized database. These records provide a comprehensive view of a patient's medical history, past diagnoses, treatments, laboratory test results, and other information, which helps clinicians understand the disease \cite{OncoEHR}. Within EHR, time-series data may refer to the clinical data recorded over time, such as repeated blood tests, lab values, or physical attributes. Such data informs the changes in the patient's condition and monitors the disease progression \cite{EHRimprove}.
    
\subsection{Taxonomy of MML} \label{Taxonomy of Multimodal Learning}
We follow the taxonomy proposed by William \emph{et al.} \cite{MultimodalClassification} (see Figure \ref{fig:fusion_stage}), which defines 5 main stages of multimodal classification: preprocessing, feature extraction, data fusion, primary learner, and final classifier, as given below:

    \subsubsection{Pre-processing} 
    Pre-processing involves modifying the input data to a suitable format before feeding it into the model for training. It includes data cleaning, normalization, class balancing, and augmentation. Data cleaning removes unwanted noise or bias, errors, and missing data points \cite{DPtxtbook}. Normalization scales the input data within a specific range to ensure that each modality contributes equally to the training \cite{ExplainingDPinML}. Class balancing is done in cases where one class may have a significantly larger number of samples than another, resulting in a model bias toward the dominant class. Data augmentation artificially increases the size of the dataset by generating new samples based on the existing data to improve the model's robustness and generalizability~\cite{DPtxtbook}.
    
    \subsubsection{Feature Extraction}
    Different data modalities may have different features, and extracting relevant features may improve model learning. Several manual and automated feature engineering techniques generate representations (or \emph{embeddings}) for each data modality. Feature engineering involves designing features relevant to the task and extracting them from the input data. This can be time-consuming but may allow the model to incorporate prior knowledge about the problem. Text encoding techniques, such as bag-of-words, word embeddings, and topic models \cite{BERT, RoBERTa}, transform textual data into a numerical representation, which can be used as input to an ML model \cite{textencoding}. In DL, feature extraction is learned automatically during model training\cite{FEconv}.   

    \subsubsection{Data Fusion}
    Data fusion combines raw features, extracted features, or class prediction vectors from multiple modalities to create a single data representation. Fusion enables the model to use the complementary information provided by each modality and improve its learning. Data fusion can be done using early, late, or intermediate fusion. Section \ref{Data Fusion Stages} discusses these fusion stages. The choice of fusion technique depends on the characteristics of the data and the specific problem being addressed \cite{jiang_sinha_aldape_hannenhalli_sahinalp_ruppin_2022}.
    
    \subsubsection{Primary Learner}
    The primary learner stage is training the model on the pre-processed data or extracted features. Depending on the problem and data, the primary learner can be implemented using various ML techniques. DNNs are a popular choice for primary learners in MML because they can automatically learn high-level representations from the input data and have demonstrated state-of-the-art performance in many applications. CNNs are often used for image and video data, while recurrent neural networks (RNNs) and Transformers are commonly used for text and sequential data. The primary learner can be implemented independently for each modality or shared between modalities, depending on the problem and data. 
    
    \subsubsection{Final Classifier}
    The final stage of MML is the classifier, which produces category labels or class scores and can be trained on the output of the primary learner or the fused data. The final classifier can be implemented using a shallow neural network, a decision tree, or an ensemble model \cite{MultimodalClassification}. Ensemble methods, such as stacking or boosting, are often used to improve and robustify the performance of the final classifier. Stacking involves training multiple models and then combining their predictions at the output stage, while boosting involves repeatedly training weak learners and adjusting their weights based on the errors made by previous learners \cite{DNNtabdatasurvey}.

\subsection{Data Fusion Strategies} \label{Data Fusion Stages}

\begin{figure}[h!]
    \centering
    \includegraphics[width=\columnwidth]{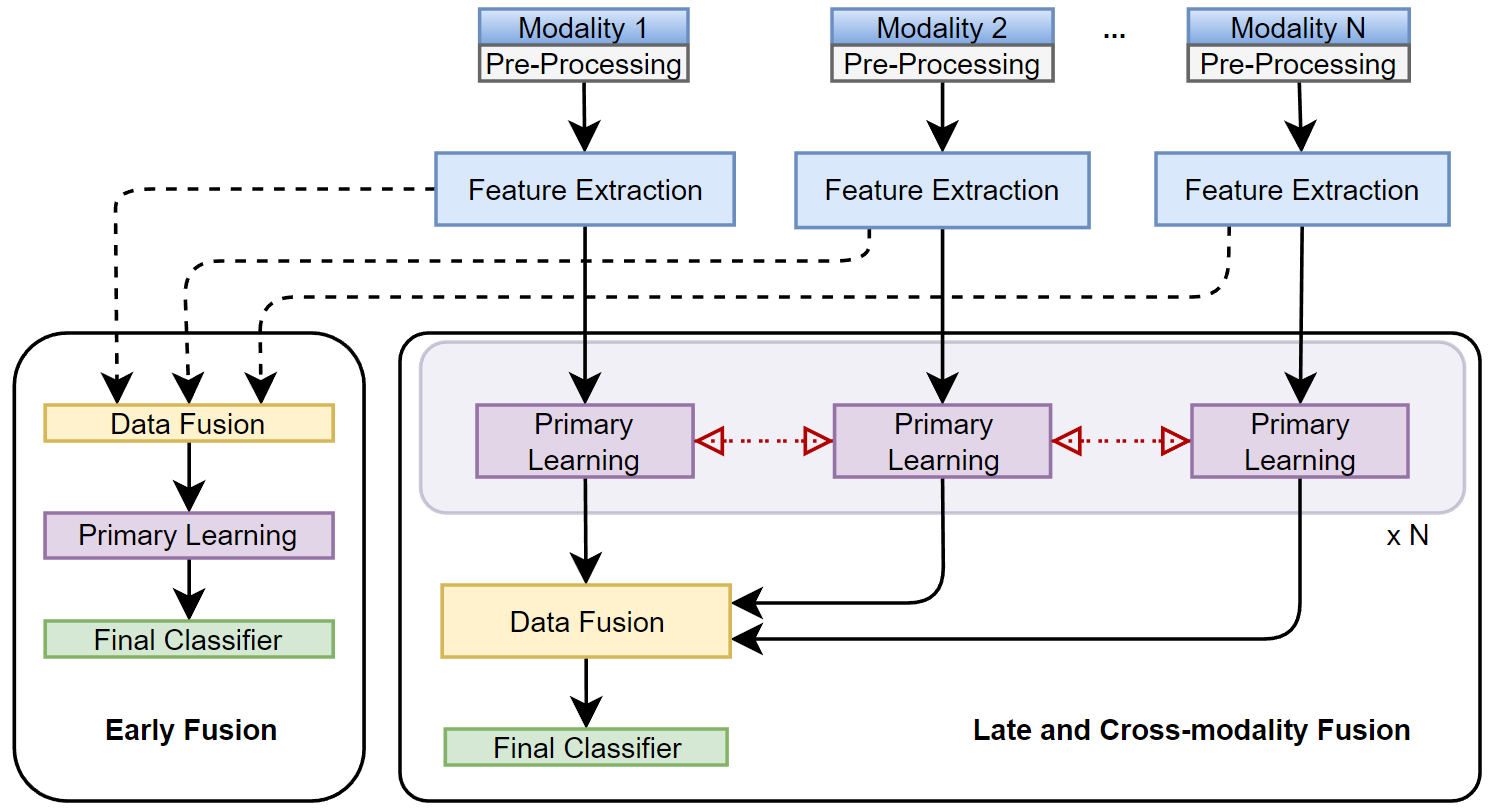}
    \caption{Taxonomy, stages, and techniques of multimodal data fusion are presented. \emph{Early}, \emph{late}, \emph{cross-modality} fusion methods integrate individual data modalities (or extracted features) \emph{before}, \emph{after}, or \emph{at} the primary learning step, respectively.}
    \label{fig:fusion_stage}
\end{figure}

Fusion in MML can be performed at different levels, including early (feature level), intermediate (model level), or late (decision level) stages, as illustrated in figure~\ref{fig:fusion_stage}. Each fusion stage has its advantages and challenges, and the choice of fusion stage depends on the characteristics of the data and the task.

\subsubsection{Early Fusion} The early fusion involves merging features extracted from different data modalities into a single feature vector before model training. The feature vectors of the different modalities are combined into a single vector, which is used as the input to the ML model \cite{MultimodalClassification}. This approach can be used when the modalities have complementary information and can be easily aligned, such as combining visual and audio features in a video analysis application. The main challenge with early fusion is ensuring that the features extracted from different modalities are compatible and provide complementary information.

\subsubsection{Intermediate Fusion}
Intermediate fusion involves training separate models for each data modality and then combining the outputs of these models for inference/prediction \cite{MultimodalClassification}. This approach is suitable when the data modalities are independent of each other and cannot be easily combined at the feature level using average, weighted average, or other methods. The main challenge with intermediate fusion is selecting an appropriate method for combining the output of different models.

\subsubsection{Late Fusion}
In late fusion, the output of each modality-specific model is used to make a decision independently. All decisions are later combined to make a final decision. This approach is suitable when the modalities provide complementary information but are not necessarily independent of each other. The main challenge with late fusion is selecting an appropriate method for combining individual predictions. This can be done using majority voting, weighted voting, or employing other ML models.

\subsection{MML for Oncology Datasets} \label{Neural Network Architectures on Oncology Data}

Syed \emph{et al.} \cite{syed2021multi} used a Random Forest classifier to fuse radiology image representations learned from the singular value decomposition method with the textual annotation representation learned from the fastText algorithm for prostate and lung cancer patients. Liu \emph{et al.} \cite{liu2022hybrid} proposed a hybrid DL framework for combining breast cancer patients' genomic and pathology data using fully-connected (FC) network for genomic data, CNN for radiology data and a Simulated Annealing algorithm for late fusion. Multiview multimodal network (MVMM-Net) \cite{song2021multiview} combined 2 different modalities (low-energy and dual-energy subtracted) from contrast-enhanced spectral mammography images, each learned through CNN and late-fusion through FC network in breast cancer detection task. Yap \emph{et al.} \cite{yap2018multimodal} used a late-fusion method to fuse image representations from ResNet50 and clinical representations from a random forest model for a multimodal skin lesion classification task. An award-winning work \cite{ma2020brain} on brain tumor grade classification adopted the late-fusion method (concatenation) for fusing outputs from two CNNs (radiology and pathology images).

The single-cell unimodal data alignment is one technique in MML. Jansen \emph{et al.} devised an approach (SOMatic) to combine ATAC-seq regions with RNA-seq genes using self-organizing maps~\cite{jansen2019building}. Single-Cell data Integration via Matching (SCIM) \cite{stark2020scim} matched cells in multiple datasets in low-dimensional latent space using autoencoder (AEs). Graph-linked unified embedding (GLUE) \cite{cao2022multi} model learned regulatory interactions across omics layers and aligned the cells using variational AEs. These aforementioned methods cannot incorporate high-order interactions among cells or different modalities. Single-cell data integration using multiple modalities is mostly based on AEs (scDART \cite{zhang2022scdart}, Cross-modal Autoencoders \cite{yang2021multi}, Mutual Information Learning for Integration of Single Cell Omics Data (SMILE) \cite{xu2022smile}).
    
\section{Graph Neural Networks (GNN\lowercase{s}) in Multimodal Learning} \label{GNNs in Multimodal Learning}
Graphs are commonly used to represent the relational connectivity of any system that has interacting entities \cite{li2022graph}. Graphs have been used in various fields, such as to study brain networks~\cite{farooq2019network}, analyze driving maps \cite{derrow2021eta}, and explore the structure of DNNs themselves \cite{waqas2022exploring}. GNNs are specifically designed to process data represented as a graph~\cite{waikhom2022survey}, which makes them well-suited for analyzing multimodal oncology data as each data modality (or sub-modality) can be considered as a single node and the structures/patterns that exist between data modalities can be modeled as edges \cite{ektefaie2023multimodal}. 
    
   \subsection{The Graph Data}
    A graph is represented as $G$=$(V, E)$ having node-set $V$=$\{v_1, v_2,..., v_n\}$, where node $v$ has feature vector \textbf{x$_{v}$}, and edge set $E$=$\{(v_i, v_j)\mid v_i, v_j\in V\}$. The neighborhood of node $v$ is defined as $N(v)$=$\{u\mid(u,v) \in E\}$. 
    
        \subsubsection{Graph Types} 
\begin{figure}[h!]
                \centering
                \includegraphics[width=\columnwidth]{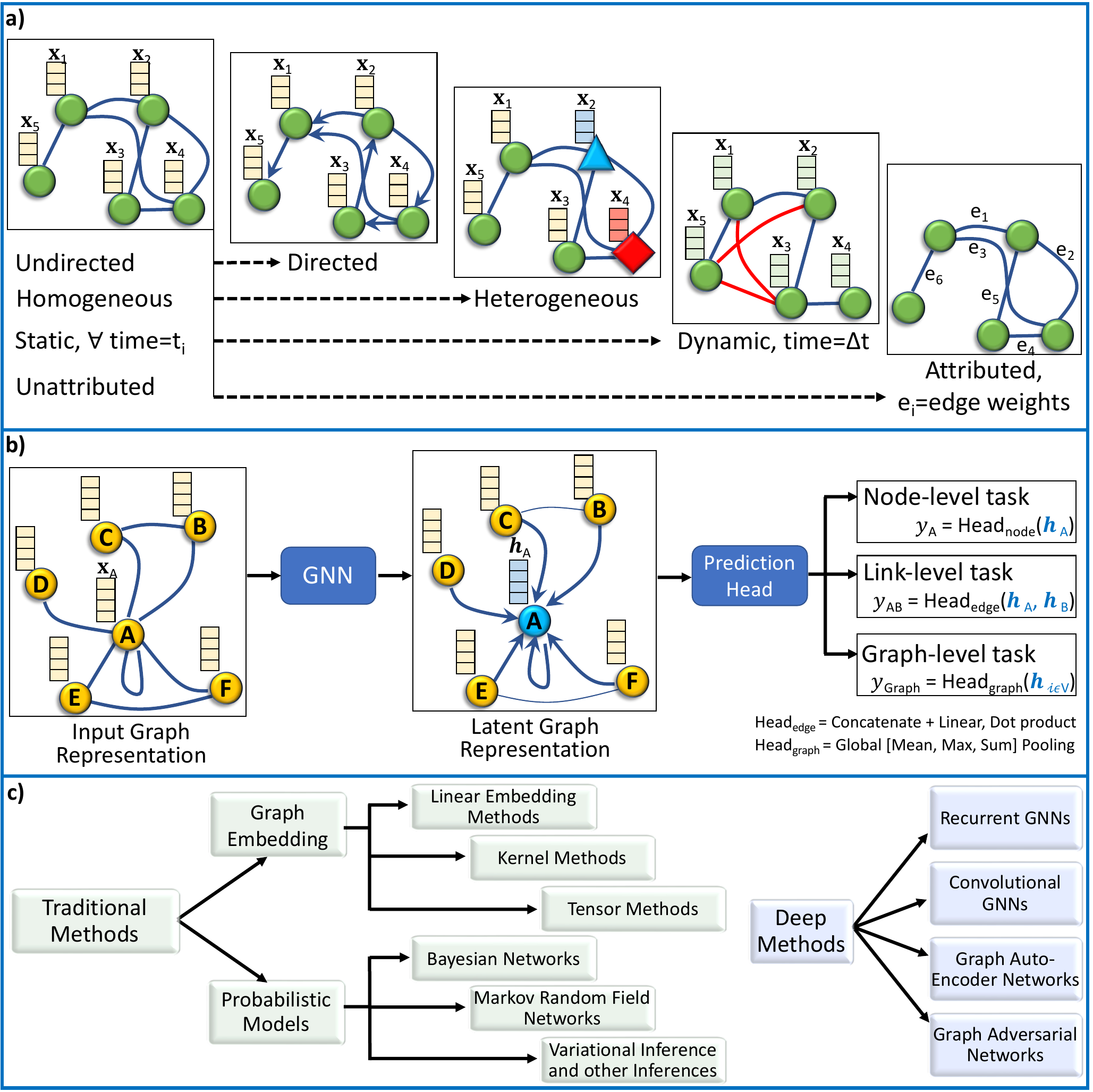}
                \caption{\textbf{(a)} The commonly occurring graph types are presented, including (1) undirected and directed, (2) homogeneous and heterogeneous, (3) dynamic and static, (4) attributed (edges) and unattributed. \textbf{(b)} Three different types of tasks performed using the graph data are presented and include (1) node-level, (2) link-level, and (3) graph-level analyses. \textbf{(c)} Various categories of representation learning for graphs are presented.}
                \label{fig:graph-combined}
            \end{figure}

As illustrated in figure \ref{fig:graph-combined}(a), the common types of graphs include undirected, directed, homogeneous, heterogeneous, static, dynamic, unattributed, and attributed.\emph{Undirected graphs} comprise undirected edges, i.e., the direction of relation is not important between any ordered pair of nodes. In the \emph{directed graphs}, the nodes have a directional relationship(s). Homogeneous graphs have the same type of nodes, whereas heterogeneous graphs have different types of nodes within a single graph \cite{yang2021hgat}. Static graphs do not change over time with respect to the existence of edges and nodes. In contrast, dynamic graphs change over time, resulting in changes in structure, attributes, and node relationships. \emph{Unattributed graphs} have unweighted edges, indicating that the weighted value for all edges in a graph is the same, i.e., $1$ if present, $0$ if absent. \emph{Attributed graphs} have different edge weights that capture the strength of relational importance \cite{waikhom2022survey}.

        \subsubsection{Tasks for Graph Data} In figure \ref{fig:graph-combined}(b), we present 3 major types of tasks defined on graphs, including (1) \emph{node-level tasks} - these may include node classification, regression, clustering, attributions, and generation, (2) \emph{edge-level task} - edge classification and prediction (presence or absence) are 2 common edge-level tasks, (3) \emph{graph-level tasks} - these tasks involve predictions on the graph level, such as graph classification and generation.

    \subsection{ML for Graph Data}
        Representing data as graphs can enable capturing and encoding the relationships among entities of the samples \cite{wu2020comprehensive}. Based on the way the nodes are encoded, representation learning on graphs can be categorized into the traditional (or shallow) and DNN-based methods, as illustrated in Figure~\ref{fig:graph-combined}(c) ~\cite{jiao2022graph, wu2020comprehensive}.
        
            \subsubsection{Traditional (Shallow) Methods} These methods usually employ classical ML methods, and their two categories commonly found in the literature are \emph{graph embedding} and \emph{probabilistic methods}. Graph embedding methods represent a graph with low-dimensional vectors (graph embedding and node embedding), preserving the structural properties of the graph. The learning tasks in graph embedding usually involve dimensionality reduction through linear (principal component or discriminant analysis), kernel (nonlinear mapping), or tensor (higher-order structures) methods~\cite{jiao2022graph}. Probabilistic graphical methods use graph data to represent probability distribution, where nodes are considered random variables, and edges depict the probability relations among nodes~\cite{jiao2022graph}. Bayesian networks, Markov's networks, variational inference, variable elimination, and others are used in probabilistic methods~\cite{jiao2022graph}.
            
            \subsubsection{DNN-based Methods - GNNs} 
\begin{figure}[htpb]
                        \centering
                        \includegraphics[width=\textwidth]{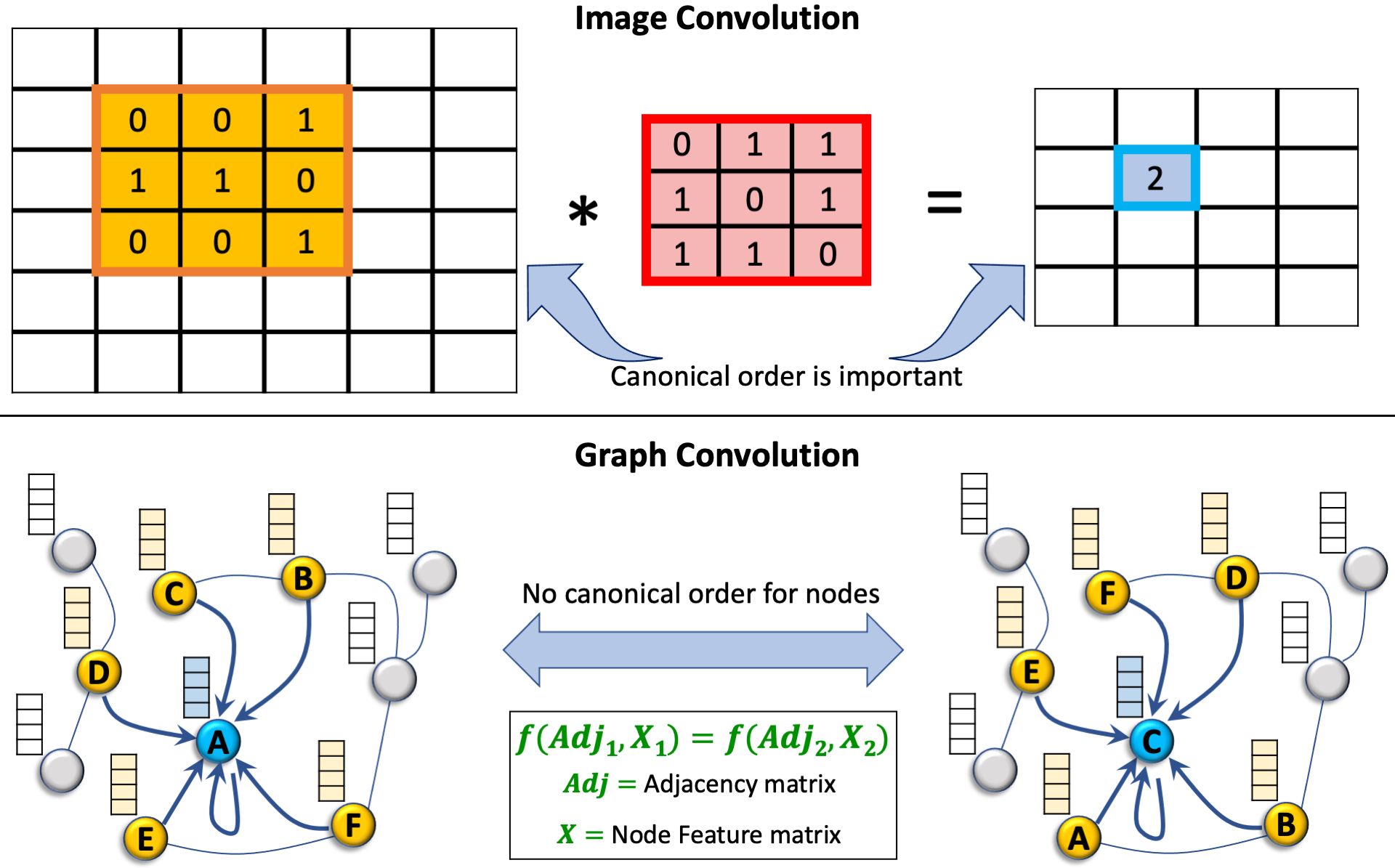}
                        \caption{Convolution operation for graphs vs. image data. The canonical order of the input is important in CNNs, whereas in GNNs, the order of the input nodes is not important. From the convolution operation perspective, CNNs can be considered a subset of GNNs \cite{hamilton2020graph}.}
                        \label{fig:convolutions}
                    \end{figure}

GNNs are gaining popularity in the ML community, as evident from figure~\ref{fig:all_publications}. In GNNs, the information aggregation from the neighborhood is fused into a node's representation. Traditional DL methods such as CNNs and their variants have shown remarkable success in processing the data in Euclidean space; however, they fail to perform well when faced with non-Euclidean or relational datasets. Compared to CNNs, where the locality of the nodes in the input is fixed, GNNs have no canonical ordering of the neighborhood of a node. They can learn the given task for any permutation of the input data, as depicted in figure \ref{fig:convolutions}. GNNs often employ a message-passing mechanism in which a node's representation is derived from its neighbors' representations via a recursive computation. The message passing for a GNN is given as follows:
                    \begin{align}
                        \mathbf{h}_{v}^{(l+1)} = & \sigma \left(W_l \sum_{u \in N(v)} \frac{\mathbf{h}_{u}^{(l)}}{|N(v)|} + B_l \mathbf{h}_{v}^{(l)}\right) 
                        \label{eq:msg-pass}
                    \end{align}
                    
            where $h_{v}^{(l+1)}$ is the updated embedding of node $v$ after $l$+$1$ layer, $\sigma$ is the non-linear function (e.g., rectified linear unit or ReLU), $h_{u}^{(l)}$ and $h_{v}^{(l)}$ represent the embeddings of nodes $u$ and $v$ at layer $l$. $W_l$ and $B_l$ are the trainable weight matrices for neighborhood aggregation and (self)hidden vector transformation, respectively. The message passing can encode high-order structural information in node embedding through multiple aggregation layers. GNNs smooth the features by aggregating neighbors’ embedding and filter eigenvalues of graph Laplacian, which provides an extra denoising mechanism \cite{ma2021unified}. GNNs comprise multiple permutation equivariant and invariant functions, and they can handle heterogeneous data \cite{jin2022heterogeneous}. As described earlier, traditional ML models deal with Euclidean data. In oncology data, the correlations may not exist in Euclidean space; instead, its features may be highly correlated in the non-Euclidean space \cite{yi2022graph}. Based on the information fusion and aggregation methodology, GNNs-based deep methods are classified into the following:
            \paragraph{Recurrent GNNs} RecGNNs are built on top of the standard Recurrent Neural Network (RNN) by combining with GNN. RecGNNs can operate on graphs with variable sizes and topologies. The recurrent component of the RecGNN captures temporal dependencies and learns latent states over time, whereas the GNN component captures the local structure. The information fusion process is repeated a fixed number of times until an equilibrium or the desired state is achieved \cite{hamilton2017inductive}. RecGNNs employ the model given by: 
                
                    \begin{equation}
                        \mathbf{h}_{v}^{(l+1)} = \text{RecNN}\left(\mathbf{h}_{u}^{(l)}, \mathbf{Msg}_{N(v)}^{(l)}\right),
                        \label{eq:RecGNN}
                    \end{equation}
                    
                where, RecNN is any RNN, and $Msg_{N(v)}^{(l)}$ is the neighborhood message-passing at layer $l$.
             \paragraph{Convolutional GNNs} ConvGNNs undertake the convolution operation on graphs by aggregating neighboring nodes' embeddings through a stack of multiple layers. ConvGNNs use the symmetric and normalized summation of the neighborhood and self-loops for updating the node embeddings given by:
                    \begin{equation}
                        \mathbf{h}_{v}^{(l+1)} = \sigma\left(W_l \sum_{u \in N(v) \cup v} \frac{\mathbf{h}_{v}}{\sqrt{|N(v)||N(u)|}}\right).
                        \label{eq:ConvGNNs}
                    \end{equation}
                The ConvGNN can be spatial or spectral, depending on the type of convolution they implement. Convolution in spatial ConvGNNs involves taking a weighted average of the neighboring vertices. Examples of spatial ConvGNNs include GraphSAGE\cite{hamilton2017inductive}, Message Passing Neural Network (MPNN)\cite{gilmer2017neural}, and Graph Attention Network (GAT)~\cite{velivckovic2017graph}. The spectral ConvGNNs operate in the spectral domain by using the eigendecomposition of the graph Laplacian matrix. The convolution operation is performed on the eigenvalues, which can be high-dimensional. Popular spectral ConvGNNs are ChebNet \cite{defferrard2016convolutional} and Graph Convolutional Network (GCN)\cite{kipf2016semi}. An interesting aspect of these approaches is representational containment, which is defined as: $\text{convolution} \subseteq  \text{attention} \subseteq \text{message passing}$.               
                
                \paragraph{Graph Auto-Encoder Networks (GAEs)}
                GAEs are unsupervised graph learning networks for dimensionality reduction, anomaly detection, and graph generation. They are built on top of the standard AEs to work with graph data. The encoder component of the GAE maps the input graph to a low-dimensional latent space, while the decoder component maps the latent space back to the original graph \cite{park2021unsupervised}.
                
                \paragraph{Graph Adversarial Networks (GraphANs)} Based on Generative Adversarial Networks, GraphANs are designed to work with graph-structured data and can learn to generate new graphs with similar properties to the input data. The generator component of the GraphAN maps a random noise vector to a new graph, while the discriminator component tries to distinguish between the generated vs. the actual input. The generator generates graphs to fool the discriminator, while the discriminator tries to classify the given graph as real or generated.
                
                \paragraph{Other GNNs} Other categories of GNNs may include scalable GNNs \cite{ma2019neugraph}, dynamic GNNs \cite{sankar2018dynamic}, hypergraph GNNs \cite{bai2021hypergraph}, heterogeneous GNNs \cite{wei2019mmgcn}, and many others~\cite{ma2021deep}. 

        \begin{figure*}[h!]
            \centering
            \includegraphics[width=\textwidth]{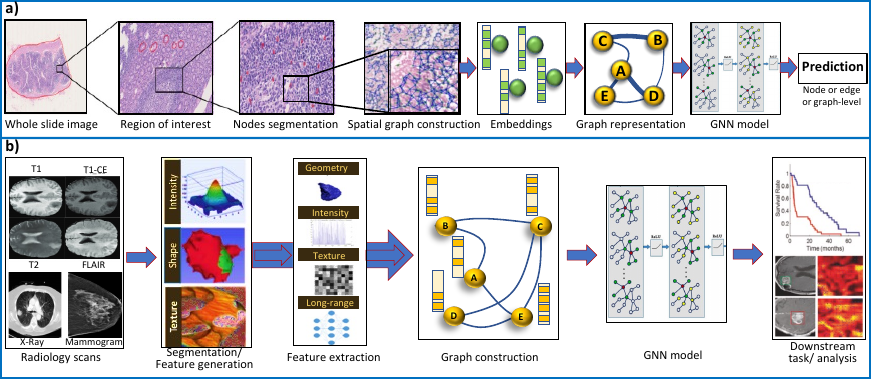}
            \caption{\textbf{(a)} Data processing pipeline for histopathology images using GNNs \cite{chen2020pathomic}. \textbf{(b)} Graph processing pipeline on radiology data. Adapted from \cite{singh2021radiomics}.}
            \label{fig:radiology-graphs}
        \end{figure*}
    
    \subsection{GNNs and ML using Unimodal Oncology Datasets}
    \subsubsection{Pathology Datasets}
    Traditionally, CNN-based models are used to learn features from digital pathology data. However, unlike GNNs, CNNs fail to capture the global contextual information important in the tissue phenotypical and structural micro and macro environment \cite{ahmedt2022survey}. For using histology images in GNNs, the cells, tissue regions, or image patches are depicted as nodes. The relations and interactions among these nodes are represented as (un)weighted edges. Usually, a graph of the patient histology slide is used along with a patient-level label for training a GNN, as illustrated in Figure~\ref{fig:radiology-graphs}(a). Here, we review a few GNN-based pathology publications representative of a trove of works in this field. Histographs \cite{anand2020histographs} used breast cancer histology data to distinguish cancerous and non-cancerous images. Pre-trained VGG-UNet was used for nuclei detection, micro-features of the nuclei were used as node features, and Euclidean distance among nuclei was incorporated as edge features. The resulting cell graphs were used to train the GCN-based robust spatial filtering (RSF) model, which performed superior to the CNN-based classification. Wang \emph{et al.} \cite{wang2020weakly} analyzed grade classification in tissue micro-arrays of prostate cancer using the weakly-supervised technique on a variant of GraphSAGE with self-attention pooling (SAGPool). Cell-Graph Signature ($CG_{signature}$)~\cite{wang2022cell} predicted patient survival in gastric cancer using cell-graphs of multiplexed immunohistochemistry images processed through two types of GNNs (GCNs and GINs) with two types of pooling (SAGPool, TopKPool). Besides the above-mentioned cell graphs, there is an elaborate review of GNN-based tissue graphs or patch-graphs methods implemented on unimodal pathology cancer data given in \cite{ahmedt2022survey}. Instead of individual cell- and tissue-graphs, a combination of the multilevel information in histology slides can help understand the intrinsic features of the disease.

    \subsubsection{Radiology Datasets}
    GNNs have been used in radiology-based cancer data for segmentation, classification, and prediction tasks, especially on X-rays, mammograms, MRI, PET, and CT scans. Figure \ref{fig:radiology-graphs}(b) illustrates a general pipeline of using radiology-based data to train GNNs. Here we give a non-exhaustive review of GNNs-based works on radiological oncology data as a single modality input. Mo \emph{et al.} \cite{mo2020multimodal} proposed a framework that improved the liver cancer lesion segmentation in the MRI-T1WI scans through guided learning of MRI-T2WI modality priors. Learned embeddings from fully convolutional networks on separate MRI modalities are projected into the graph domain for learning by GCNs through the co-attention mechanism and finally to get the refined segmentation by re-projection. Radiologists usually review radiology images by zooming into the region of interest (ROIs) on high-resolution monitors. Du \emph{et al.} \cite{du2019zoom} used a hierarchical GNN framework to automatically zoom into the abnormal lesion region of the mammograms and classify breast cancer. The pre-trained CNN model extracts image features, whereas a GAT model is used to classify the nodes for deciding whether to zoom in or not based on whether it is benign or malignant. Based on the established knowledge that lymph nodes (LNs) have connected lymphatic system and LNs cancer cells spread on certain pathways, Chao \emph{et al.}~\cite{chao2020lymph} proposed a lymph node gross tumor volume learning framework. The framework was able to delineate the LN appearance as well as the inter-LN relationship. The end-to-end learning framework was superior to the state-of-the-art on esophageal cancer radiotherapy dataset. Tian \emph{et al.} \cite{tian2020graph} suggested interactive segmentation of MRI scans of prostate cancer patients through a combination of CNN and two GCNs. CNN model outputs a segmentation feature map of MRI, and the GCNs predict the prostate contour from this feature map. Saueressig {et al.} \cite{saueressig2021exploring} used GNNs to segment brain tumors in 3D MRI images, formed by stacking different modalities of MRI (T1, T2, T1-CE, FLAIR) and representing them as supervoxel graph. The authors reported that GraphSAGE-pool was best for segmenting brain tumors. Besides radiology, a parallel field of radiomics has recently gained attraction. Radiomics is the automated extraction of quantitative features from radiology scans. A survey of radiomics and radiogenomic analysis on brain tumors is presented in\cite{singh2021radiomics}. 
    
    \subsubsection{Molecular Datasets}
\begin{figure}[h!]
                \centering
                \includegraphics[width=\textwidth]{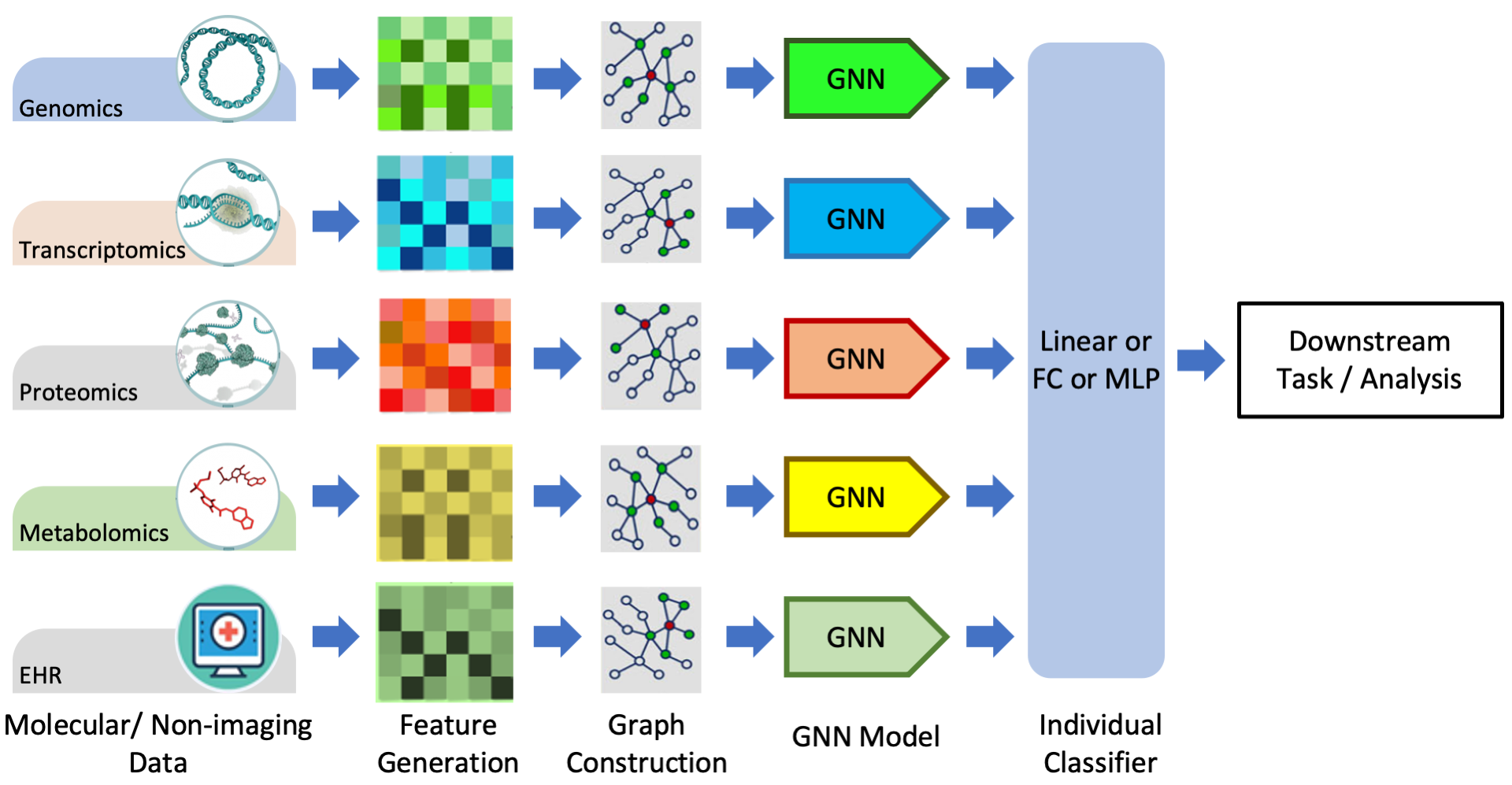}
                \caption{Graph data processing pipeline for non-imagery data, including molecular and textual data. Adapted from \cite{wang2021mogonet}. Abbreviations used: GNN - graph neural network, FC - Fully-Connected, MLP - Multi-Layer Perception.}
                \label{fig:molecular-graphs}
        \end{figure}

     Graphs are a natural choice for representing molecular data such as omic-centric (DNA, RNA, or proteins) or single-cell centric. Individual modalities are processed separately to generate graph representations that are then processed through GNNs followed by the classifier to predict the downstream task, as illustrated in Figure \ref{fig:molecular-graphs}. One method of representing proteins as graphs is to depict the amino acid residue in the protein as the node and the relationship between residues denoted by edge~\cite{fout2017protein}. The residue information is depicted as node embedding, whereas the relational information between two residues is represented as the edge feature vector. Fout \emph{et al.} \cite{fout2017protein} used spatial ConvGNNs to predict interfaces between proteins which is important in drug discovery problems. Deep predictor of drug-drug interactions (DPDDI) predicted the drug-drug interactions using GCN followed by a 5-layer classical neural network \cite{feng2020dpddi}. Molecular pre-training graph net (MPG) \cite{li2021effective} is a powerful framework based on GNN and Bidirectional Encoder Representations from Transformers (BERT) to learn drug-drug and drug-target interactions. Graph-based Attention Model (GRAM) \cite{choi2017gram} handled the data inefficiency by supplementing EHRs with hierarchical knowledge in the medical ontology. A few recent works have applied GNNs to single-cell data. scGCN \cite{song2021scgcn} is a knowledge transfer framework in single-cell omics data such as mRNA or DNA. scGNN \cite{wang2021scgnn} processed cell-cell relations through GNNs for the task of missing-data imputation and cell clustering on single-cell RNA sequencing (scRNA-seq) data.
            
    \subsection{MML - Data Fusion at the Pre-Learning Stage} \label{Multimodal Pre-learning Fusion}
    The first and most primitive form of MML is the pre-learning fusion (see Figure \ref{fig:fusion_stage}), where features extracted from individual modalities of data are merged, and the fused representations are then used for training the multimodal primary learner model. In the context of GNNs being the primary learning model, the extraction step of individual modality representations can be hand-engineered (e.g., dimensionality reduction) or learned by DL models (e.g., CNNs, Transformers). Cui \emph{et al.} \cite{cui2021co} proposed a GNN-based early fusion framework to learn latent representations from radiological and clinical modalities for Lymph node metastasis (LNM) prediction in esophageal squamous cell carcinoma (ESCC). The extracted features from the two modalities using UNet and CNN-based encoders were fused together with category-wise attention as node representation. The message passing from conventional GAT and correlation-based GAT learned the neighborhood weights. The attention attributes were used to update the final node features before classification by a 3-layer fully connected network. For Autism spectrum disorder, Alzheimer’s disease, and ocular diseases, a multimodal learning framework called Edge-Variational GCN (EV-GCN) \cite{huang2020edge} fuses the radiology features extracted from fMRI images with clinical feature vectors for each patient. An MLP-based pairwise association encoder is used to fuse the input feature vectors and to generate the edge weights of the population graph. The partially labeled population graph is then processed through GCN layers to generate the diagnostic graph of patients.
    
    \subsection{MML - Data Fusion using Cross-Modality Learning} \label{Multimodal Cross-learning Fusion}
    Cross-MML involves intermediate fusion and/or cross-links among the models being trained on individual modalities (see Figure~\ref{fig:fusion_stage}). For this survey, we consider the GNN-based hierarchical learning mechanisms as the cross-MML methods. Hierarchical frameworks involve learning for one modality and using the learned latent embeddings in tandem with other data modalities sequentially to get the final desired low-dimensional representations. Lian \emph{et al.}~\cite{NSCLCGNNTransformer} used a sequential learning framework where tumor features learned from CT images using the ViT model were used as node features of the patient population graph for subsequent processing by the GraphSAGE model. The hierarchical learning from radiological and clinical data using Transformer-GNN outperformed the ResNet-Graph framework in survival prediction of early-stage NSCLC. scMoGNN \cite{wen2022graph} is the first method to apply GNNs in multimodal single-cell data integration using a cross-learning fusion-based GNN framework. Officially winning first place in modality prediction task at the NeurIPS 2021 competition, scMoGNN showed superior performance on various tasks by using paired data to generate cell-feature graphs. Hierarchical cell-to-tissue-graph network (HACT-Net) combined the low-level cell-graph features with the high-level tissue-graph features through two hierarchical GINs on breast cancer multi-class prediction \cite{pati2020hact}. Data imputation, a method of populating the missing values or false zero counts in single-cell data mostly done using DL autoencoders (AE) architecture, has recently been accomplished using GNNs. scGNN~\cite{wang2021scgnn} used imputation AE and graph AE in an iterative manner for imputation, and GraphSCI \cite{rao2021imputing} used GCN with AE to impute the single-cell RNA-seq data using the cross-learning fusion between the GCN and the AE networks. Clustering is a method of characterizing cell types within a tissue sample. Graph-SCC~\cite{zeng2020accurately} clustered cells based on scRNA-seq data through self-supervised cross-learning between GCN and a denoising AE network. Recently, a multilayer GNN framework, Explainable Multilayer GNN (EMGNN), has been proposed for cancer gene prediction tasks using multi-omics data from 16 different cancer types \cite{chatzianastasis2023explainable}. 
    
    \subsection{MML - Data Fusion in Post-Learning Regime} \label{Multimodal Post-learning}
    Post-learning fusion methods include processing individual data modalities and later fusing them for the downstream predictive task~\cite{tortora2022radiopathomics}. In the post-learning fusion paradigm, the hand-crafted features perform better than the deep features when the dimensionality of input data is low, and vice versa~\cite{tortora2022radiopathomics}. Many interesting GNN-based works involving the post-learning fusion mechanism have recently been published. Decagon~\cite{zitnik2018modeling} used a multimodal approach on GCNs using proteins and drug interactions to predict exact side effects as a multi-relational link prediction task. Drug–target affinity (DTA)~\cite{nguyen2021graphdta} experimented with four different flavors of GNNs (GCN, GAT, GIN, GAT-GCN) along with a CNN to fuse together molecular embeddings and protein sequences for predicting drug-target affinity. PathomicFusion~\cite{chen2020pathomic} combined the morphological features extracted from image patches (using CNNs), cell-graph features from cell-graphs of histology images (GraphSAGE-based GCNs), and genomic features (using a feed-forward network) for survival prediction on glioma and clear cell renal cell carcinoma. Shi \emph{et al.}~\cite{shi2019graph} proposed a late-fusion technique to study screening of cervical cancer at early stages by using CNNs to extract features from histology images, followed by K-means clustering to generate graphs which are processed through two-layer GCN. BDR-CNN-GCN (batch normalized, dropout, rank-based pooling)~\cite{zhang2021improved} used the same mammographic images to extract image-level features using CNN and relation-aware features using GCN. The two feature sets are fused using a dot product followed by a trainable linear projection for breast cancer classification. Under the umbrella of multi-omics data, many GNN-based frameworks have been proposed recently. Molecular omics network(MOOMIN)\cite{rozemberczki2022moomin}, a multi-modal heterogeneous GNN to predict oncology drug combinations, processed molecular structure, protein features, and cell lines through GCN-based encoders, followed by late-fusion using a bipartite drug-protein interaction graph. Multi-omics graph convolutional networks (MOGONET) \cite{wang2021mogonet} used a GCN-GAN late fusion technique for the classification of four different diseases, including three cancer types: breast, kidney, and glioma. Leng \emph{et al.} \cite{leng2022benchmark} extended MOGONET to benchmark three multi-omics datasets on two different tasks using sixteen DL networks and concluded that GAT-based GNN had the best classification performance. Multi-Omics Graph Contrastive Learner(MOGCL) \cite{rajadhyaksha2023graph} used graph structure and contrastive learning information to generate representations for improved downstream classification tasks on the breast cancer multi-omics dataset using late-fusion. Similar to MOGCL, Park \emph{et al.} \cite{park2022deep} developed a GNN-based multi-omics model that integrated mRNA expression, DNA methylation, and DNA sequencing data for NSCLC diagnosis. 
    
\section{Transformer\lowercase{s} in MML} \label{Transformers in Multi-modal Learning}
Transformers are attention-based DNN models originally proposed for NLP \cite{vaswani2017attention}. Transformers implement scaled dot-product of the input with itself and can process various types of data in parallel \cite{vaswani2017attention}. Transformers can handle sequential data and learn long-range dependencies, making them well-suited for tasks such as language translation, language modeling, question answering, and many more \cite{surveyonNLPtransformers}. Unlike Recurrent Neural Networks (RNNs) and CNNs, Transformers use self-attention operations to weigh the importance of different input tokens (or embeddings) at each time step. This allows them to handle sequences of arbitrary length and to capture dependencies between input tokens that are far apart in the sequence \cite{vaswani2017attention}. Transformers can be viewed as a type of GNN~\cite{MLwithTransformersAsurvey}. Transformers are used to process other data types, such as images~\cite{ViT}, audio \cite{transformersAudio}, and time-series analysis \cite{ahmed2022transformers}, resulting in a new wave of multi-modal applications. Transformers can handle input sequences of different modalities in a unified way, using the same self-attention mechanism, which processes the inputs as a fully connected graph \cite{MLwithTransformersAsurvey}. This allows Transformers to capture complex dependencies between different modalities, such as visual and textual information in visual question-answering (VQA) tasks \cite{multitaskVQA}. 

Pre-training Transformers on large amounts of data, using unsupervised or self-supervised learning, and then fine-tuning for specific downstream tasks, has led to the development of foundation models \cite{boehm2021harnessing}, such as BERT \cite{BERT}, GPT~\cite{GPT}, RoBERTa~\cite{RoBERTa}, CLIP~\cite{CLIP}, T5~\cite{T5}, BART \cite{BART}, BLOOM~\cite{BLOOM}, ALIGN~\cite{ALIGN}, CoCa~\cite{CoCa} and more. Multimodal Transformers are a recent development in the field of MML, which extends the capabilities of traditional Transformers to handle multiple data modalities. The inter-modality dependencies are captured by the cross-attention mechanism in multimodal Transformers, allowing the model to jointly reason and extract rich data representations. There are various types of multimodal Transformers, such as Unified Transformer (UniT) \cite{UniT}, Multi-way Multimodal Transformer (MMT) \cite{MMT}, CLIP \cite{CLIP}, Flamingo \cite{Flamingo}, CoCa \cite{CoCa}, Perceiver IO \cite{PerceiverIO}, and GPT-4\cite{openai2023gpt4}.

    \begin{figure}[h!]
        \centering
        \includegraphics[width=\textwidth]{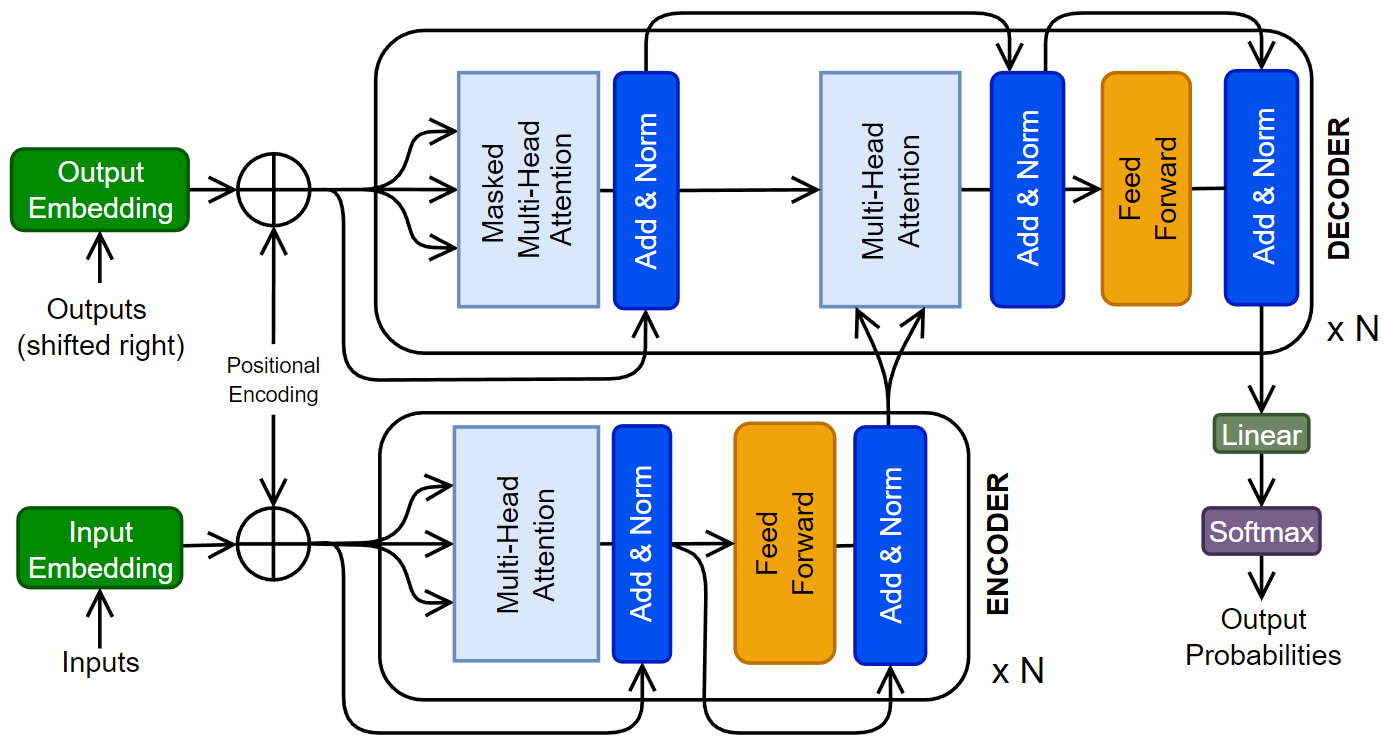}
        \caption{The original Transformer architecture is presented \cite{vaswani2017attention}. A Transformer can have multiple encoder and decoder blocks, as well as some additional layers.}
        \label{fig:vanillaTransformer}
    \end{figure}

\subsection{Model Architecture}
    
    The original Transformer (Figure \ref{fig:vanillaTransformer}) was composed of multiple encoder and decoder blocks, each made up of several layers of self-attention and feed-forward neural networks. The encoder takes the input sequence and generates hidden representations, which are then fed to the decoder. The decoder generates the output sequence by attending to the encoder's hidden representations and the previous tokens (i.e., auto-regressive). The self-attention operation (or scaled dot-product) is a crucial component of the Transformer. It determines the significance of each element in the input sequence with respect to the whole input. Self-attention operates by computing a weighted sum of the input sequence's hidden representations, where the weights are determined by the dot product between the \emph{query} vector and the \emph{key} vector, followed by a scaling operation to stabilize the gradients. The resulting weighted sum is multiplied by a \emph{value} vector to obtain the output of the self-attention operation. There has been a tremendous amount of work on various facets of Transformer architecture. The readers are referred to relevant review papers \cite{surveyonNLPtransformers, MLwithTransformersAsurvey, surveyViT, attentioninNLP}.

    \subsection{Multimodal Transformers}
\begin{figure}[h!]
            \centering
            \includegraphics[width=\textwidth]{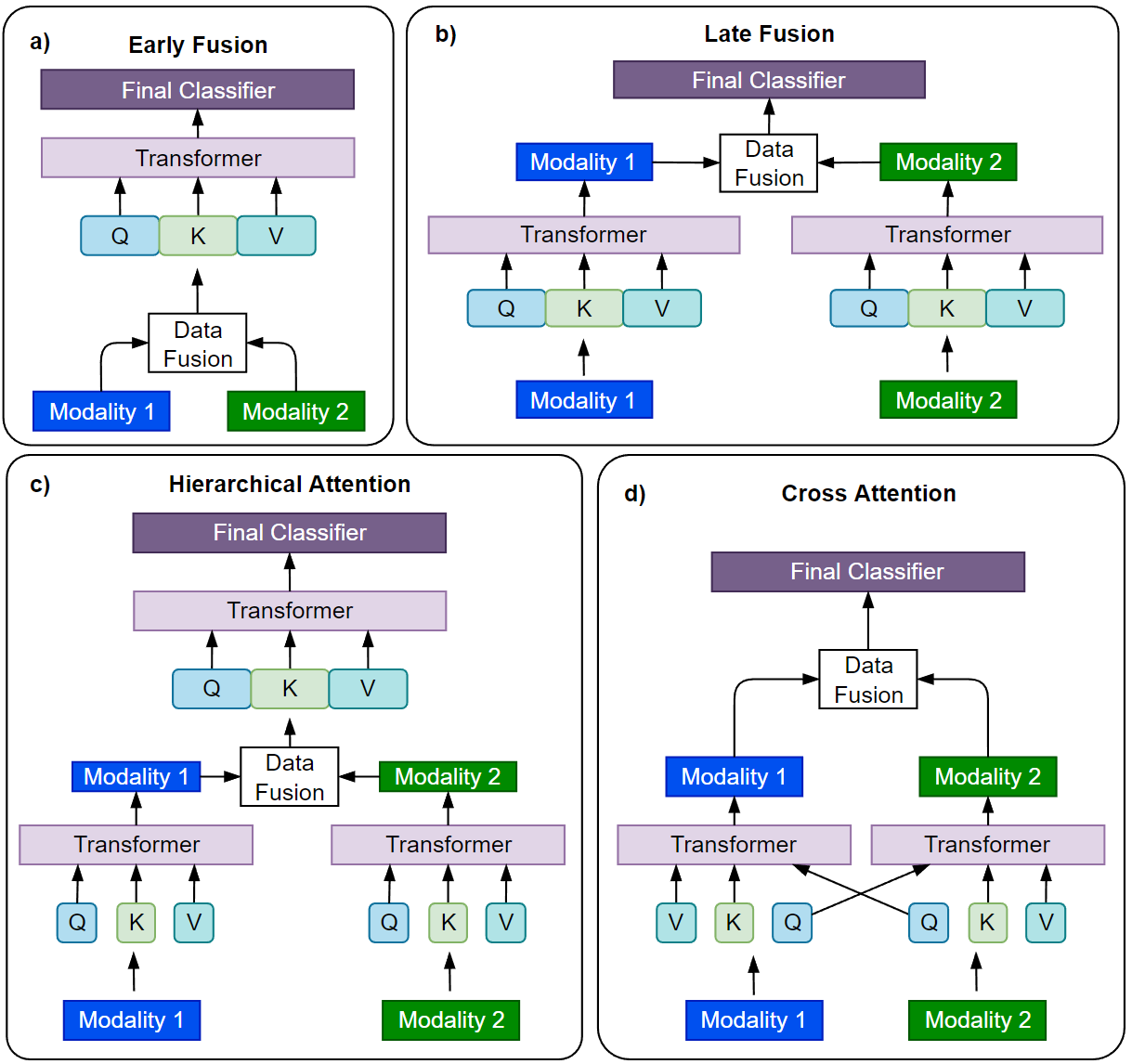}
            \caption{Four different strategies of fusing information from various data modalities in multimodal Transformers are presented.}
            \label{fig:MMLtransformerinteractions}
        \end{figure}

    Self-attention allows a Transformer model to process each input as a fully connected graph and attend to (or equivalently learn from) the global patterns present in the input. This makes Transformers compatible with various data modalities by treating each token (or its embedding) as a node in the graph. To use Transformers for a data modality, we need to tokenize the input and select an embedding space for the tokens. Tokenization and embedding selections are flexible and can be done at multiple granularity levels, such as using raw features, ML-extracted features, patches from the input image, or graph nodes. Table \ref{tab1} summarizes some common practices used for various data types in oncology datasets. Handling inter-modality interactions is the main challenge in developing multimodal Transformer models. Usually, it is done through one of these fusion methods: \emph{early fusion} of data modalities, \emph{cross-attention}, \emph{hierarchical attention}, and \emph{late fusion}, as illustrated in Figure~\ref{fig:MMLtransformerinteractions}. In the following, we present and compare data processing steps for these four methods using two data modalities as an example. The same analysis can be extended to multiple modalities.

\begin{table}
    \begin{center}
    \caption{Oncology data modalities and their respective tokenization and embeddings selection techniques}
    \label{tab1}
        \begin{tabular}{l|l|l}
            \hline \hline
            \textbf{Data Modalities} & \textbf{Tokenization Level} & \textbf{Token Embeddings Model} \\ \hline \hline
            Pathology images & Patch & CNNs \cite{MCAT} \\ \hline
            Radiology images & Patch & CNNs \cite{CoTr} \\ \hline
            EHR data & ICD code & \begin{tabular}[c]{@{}l@{}}GNNs \cite{G-BERT},\\ ML models \cite{Med-BERT} \end{tabular} \\ \hline
            -Omics & \begin{tabular}[c]{@{}l@{}}Graphs\\ K-mers\end{tabular} & \begin{tabular}[c]{@{}l@{}}GNNs \cite{MOGT}\\ ML model \cite{dnaBERT}\end{tabular} \\ \hline
            Clinical notes & Word & \begin{tabular}[c]{@{}l@{}}BERT \cite{BERT},  \\ RoBERTa \cite{RoBERTa}, \\ BioBERT \cite{BioBERT}\end{tabular} \\ \hline \hline
        \end{tabular}
    \end{center}
\end{table}

    \subsubsection{Early Fusion}
    Early fusion is the simplest way to combine data from multiple modalities. The data from different modalities are concatenated to a single input before being fed to the Transformer model, which processes the input as a single entity. Mathematically, the concatenation operation is represented as $x_{cat}$=$[x_1, x_2]$, where $x_1$ and $x_2$ are the inputs from two data modalities, and $x_{cat}$ is the concatenated input to the model. Early fusion is simple and efficient. However, it assumes that all modalities are equally important and relevant for the task at hand \cite{LatewithBERT}, which may not always be practically true \cite{FFTransformerforMM}. 
    
    \subsubsection{Cross-Attention Fusion}
    Cross-attention is a relatively more flexible approach to modeling the interactions between data modalities and learning their joint representations. The self-attention layers attend to different modalities at different stages of data processing. Cross-attention allows the model to selectively attend to different modalities based on their relevance to the task \cite{SelfDoc} and capture complex interactions between the modalities \cite{stableDiffusion}.
    
    \subsubsection{Hierarchical Fusion}
    Hierarchical fusion is a complex approach to combining multiple modalities. For instance, the Depth-supervised Fusion Transformer for Salient Object Detection (DFTR) employs hierarchical feature extraction to improve salient object detection performance by fusing low-level spatial features and high-level semantic features from different scales \cite{DFTR}. Yang \emph{et al.}\cite{HTML} introduced a hierarchical approach to fine-grained classification using a fusion Transformer. Furthermore, the Hierarchical Multimodal Transformer (HMT) for video summarization can capture global dependencies and multi-hop relationships among video frames \cite{HierarchicalVideo}. 
    
    \subsubsection{Late Fusion} 
    In late fusion, each data modality is processed independently by its own Transformer model, the branch outputs are concatenated and passed through the final classifier. Late fusion allows the model to capture the unique features of each modality while still learning their joint representation. Sun \emph{et al.} proposed a multi-modal adaptive late fusion Transformer network for estimating the levels of depression \cite{MMdepression}. Their model extracts long-term temporal information from audio and visual data independently and then fuses weights at the end to learn a joint representation of data.

    \subsection{Transformers for Processing Oncology Datasets}
    Transformers have been successfully applied to various tasks in oncology, including cancer screening, diagnosis, prognosis, treatment selection, and prediction of clinical variables \cite{boehm2021harnessing, shao2021transmil, TransConver, NSCLCGNNTransformer, MCAT}. For instance, a Transformer-based model was used to predict the presence and grade of breast cancer using a combination of imaging and genomics data \cite{boehm2021harnessing}. TransMIL \cite{shao2021transmil}, a Transformer model, was proposed to process histopathology images using self-attention to learn and classify histopathology slides by overcoming the challenges faced by multi-instance learning (MIL). Recently, a Transformer and convolution parallel network, TransConv \cite{TransConver}, was proposed for automatic brain tumor segmentation using MRI data. Transformers and GNNs have also been combined in MML for early-stage NSCLC prognostic prediction using the patient's clinical and pathological features and by modeling the patient's physiological network \cite{NSCLCGNNTransformer}. Similarly, a multimodal co-attention Transformer was proposed for survival prediction using WSIs and genomic sequences \cite{MCAT}. The authors used a co-attention mechanism to learn the interactions between the two data modalities.

\section{MML - Challenges and Opportunities} \label{Challenges and Opportunities}

Learning from multimodal oncology data is a complex and rapidly growing field that presents both challenges and opportunities. While MML has shown significant promise, there are many challenges owing to the inductive biases of the ML models \cite{ektefaie2022geometric}. In this context, we present major challenges of MML in oncology settings that, if addressed, could unlock the full potential of this emerging field. 
  
    \subsection{Large Amounts of High-quality Data}
    DL models are traditionally trained on large datasets with enough samples for training, validation, and testing, such as JFT-300M \cite{sun2017revisiting} and YFCC100M \cite{thomee2016yfcc100m}, which are not available in the cancer domain. For example, the largest genomics data repository, the Gene Expression Omnibus (GEO) database, has approximately 1.1 million samples with the keyword `cancer’ compared to 3 billion images in JFT-300M \cite{jiang2022big}. Annotating medical and oncology data is a time-consuming and manual process that requires significant expertise in many different areas of medical sciences. Factors like heterogeneity of the disease, noise in data recording, background, and training of medical professionals leading to inter- and intra-operator variability cause lack of reproducibility and inconsistent clinical outcomes \cite{MMLdatainteinonco}.
    
    \subsection{Data Registration and Alignment}
    Data alignment and registration refer to the process of combining and aligning data from different modalities in a useful manner \cite{zhao2023accurate}. In multimodal oncology data, this process involves aligning data from multiple modalities such as CT, MRI, PET, and WSIs, as well as genomics, transcriptomics, and clinical records. Data registration involves aligning the data modalities to a common reference frame and may involve identifying common landmarks or fiducial markers. If the data is not registered or aligned correctly, it may be difficult to fuse the information from different modalities accurately \cite{liang2022foundations}.
    
    \subsection{Pan-Cancer Generalization and Transference}
    Transference in MML aims to transfer knowledge between modalities and their representations to improve the performance of a model trained on a primary modality \cite{liang2022foundations}. Because of the unique characteristics of each cancer type and site, it is challenging to develop models that can generalize across different cancer sites. Furthermore, the models trained on a specific modality, such as radiology images, will not perform well with other imaging modalities, such as histopathology slides. Fine-tuning the model on a secondary modality, multimodal co-learning, and model induction are techniques to achieve transference and generalization \cite{wei2020combating}. To overcome this challenge, mechanisms for improved universality of ML models need to be devised. 

    \subsection{Missing Data Samples and Modalities}
    The unavailability of one or more modalities or the absence of samples in a modality affects the model learning, as most of the existing DL models cannot process the ``missing information''. This requirement, in turn, constrains the already insufficient size of datasets in oncology. Almost all publicly available oncology datasets have missing data for a large number of samples \cite{jiang2022big}. Various approaches for handling missing data samples and modalities are gradually gaining traction. However, this is still an open challenge \cite{mirza2019machine}. 
    
    \subsection{Imbalanced Data}
    Class imbalance refers to the phenomenon when one class (e.g., cancer negative/positive) is represented significantly more in the data than another class. Class imbalance is common in oncology data \cite{mirza2019machine}. DL models struggle to classify underrepresented classes accurately. Techniques such as data augmentation, ensemble, continual learning, and transfer learning are used to counter the class imbalance challenge \cite{mirza2019machine}. 
    
    \subsection{Explainability and Trustworthiness}
    The explainability in DL, e.g., how GNNs and Transformers make a specific decision, is still an area of active research \cite{li2022explainability, nielsen2022robust}. GNNExplainer \cite{ying2019gnnexplainer}, PGM-Explainer \cite{vu2020pgm}, and SubgraphX \cite{yuan2021explainability} are some attempts to explain the decision-making process of GNNs. The explainability methods for Transformers have been analyzed in \cite{remmer2022explainability}. Existing efforts and a roadmap to improve the trustworthiness of GNNs have been presented in the latest survey \cite{zhang2022trustworthy}. However, the explainability and trustworthiness of multimodal GNNs and Transformers is an open challenge. 
    
    \subsection{Over-smoothing in GNNs}
    One particular challenge in using GNNs is over-smoothing, which occurs when the GNN is trained for too long, causing the node representations to become almost similar \cite{wu2020comprehensive}. This leads to a loss of information, a decrease in the model's performance, and a lack of generalization \cite{valsesia2021ran}. Regularization techniques such as dropout, weight decay, skip-connection, and incorporating higher-order structures, such as motifs and graphlets, have been proposed. However, building deep architectures that can scale and adapt to varying structural patterns of graphs is still an open challenge. 
    
    \subsection{Modality Collapse}
    Modality collapse is a phenomenon that occurs in MML, where a model trained on multiple modalities may become over-reliant on a single modality, to the point where it ignores or neglects the other modalities \cite{javaloy2022mitigating}. Recent work explored the reasons and theoretical understanding of modality collapse \cite{huang2022modality}. However, the counter-actions needed to balance model dependence on data modalities require active investigation by the ML community. 
    
    \subsection{Dynamic and Temporal Data}
    Dynamic and temporal data refers to the data that changes over time \cite{wu2020comprehensive}. Tumor surveillance is a well-known technique to study longitudinal cancer growth over multiple data modalities \cite{waqas2021brain}. Spatio-temporal methods such as multiple instance learning, GNNs, and hybrid of multiple models can capture complex change in the data relationships over time; however, learning from multimodal dynamic data is very challenging and an active area of research \cite{fritz2022combining}.
    
    \subsection{Data Privacy and Federated Learning}
    With the increased concern for the privacy of data, especially in medical settings, MML techniques need to adapt to local data processing and remote federation. Federated learning can help train large multimodal models across various sites without sharing data \cite{pati2022federated}.
    
    \subsection{Other Challenges}
    MML requires extensive computational resources to train models on a variety of datasets and tasks. Robustness and failure detection \cite{ahmed2022failure} are critical aspects of MML, particularly in applications such as oncology. Uncertainty quantification techniques, such as Bayesian neural networks \cite{dera2021premium}, are still under-explored avenues in the MML. By addressing these challenges, it is possible to develop MML models that are able to surpass the performance offered by single-modality models.
    
\section{Multimodal Oncology Data Sources} \label{Multimodal Oncology Data Sources}
Building central archives to unify the different collections of oncology data requires a dedicated effort. We have compiled a non-exhaustive list of datasets from data portals shared by the National Institute of Health and others. The purpose of this compilation is to provide ML oncology researchers with a unified source of data. This periodically updated collection is available at \url{https://lab-rasool.github.io/pan-cancer-dataset-sources/}, \cite{tripathi2024building}.

\section{Conclusion}
Existing research into the integration of data across various modalities has already yielded promising outcomes, highlighting the potential for significant advancements in cancer research. However, the lack of a comprehensive framework capable of encompassing the full spectrum of cancer dataset modalities presents a notable challenge. The synergy between diverse methodologies and data across different scales could unlock deeper insights into cancer, potentially leading to more accurate prognostic and predictive models than what is possible through single data modalities alone. In our survey, we have explored the landscape of multimodal learning applied to oncology datasets and the specific tasks they can address. Looking ahead, the key to advancing this field lies in the development of robust, deployment-ready MML frameworks. These frameworks must not only scale efficiently across all modalities of cancer data but also incorporate capabilities for uncertainty quantification, interpretability, and generalizability. Such advancements will be critical for effectively integrating oncology data across multiple scales, modalities, and resolutions. The journey towards achieving these goals is complex, yet essential for the next leaps in cancer research. By focusing on these areas, future research has the potential to significantly enhance our understanding of cancer, leading to improved outcomes for patients through more informed and personalized treatment strategies.

\section{Acknowledgments}
\noindent This work was partly supported by the National Science Foundation awards ECCS-1903466, OAC-2008690 and OAC-2234836.

\bibliographystyle{IEEEtran}
\bibliography{thebibliography}

\end{document}